%% file: eacl2023.tex
\pdfoutput=1

\documentclass[11pt]{article}

\usepackage[]{EACL2023}

\usepackage{times}
\usepackage{latexsym}

\usepackage[T1]{fontenc}

\usepackage[utf8]{inputenc}

\usepackage{microtype}
\usepackage{microtype}
\usepackage[utf8]{inputenc}
\usepackage{times}
\usepackage{latexsym}
\usepackage{times}
\usepackage{latexsym}
\usepackage{multicol}
\usepackage{multirow}
\usepackage{graphicx}
\usepackage{booktabs}
\usepackage{microtype}
\usepackage{algorithm}
\usepackage{algorithmic}
\usepackage{amsbsy}
\usepackage{amsmath}
\usepackage{multirow}
\usepackage{makecell}

\usepackage{amsmath,soul}
\usepackage{soulpos}
\usepackage{tablefootnote}

\usepackage{amssymb}

\newcommand{\stitle}[1]{\vspace{0.3ex} \noindent{\bf #1}}

\usepackage{tikz}
\usepackage{hhline}
\usepackage[clock]{ifsym}
\usepackage{fontawesome}

\title{Generating Synthetic Speech from \textit{SpokenVocab}  for Speech Translation}

\author{Jinming Zhao\ \ \ \ \ \ \ \ \ \ \ \ \ \ \ Gholamreza Haffari\ \ \ \ \ \ \ \ \ \ \ \ \ \ \ Ehsan Shareghi \\
Department of Data Science \& AI, Monash University \\
\texttt{firstname.lastname@monash.edu}}

\begin{document}
\maketitle

\input{sections/0-abstract}

\input{sections/1-introduction}

\input{sections/2-method}

\input{sections/3-experiment}

\input{sections/4-results}

\input{sections/5-conclusion}

\input{sections/limitation}

\bibliography{anthology,custom}
\bibliographystyle{acl_natbib}

\clearpage
\appendix
\section{Appendix}
\input{sections/6-appendix}


\end{document}

%% file: sections/0-abstract.tex
\begin{abstract}

Training end-to-end speech translation (ST) systems requires sufficiently large-scale data, which is unavailable for most language pairs and domains. One practical solution to the data scarcity issue is to convert  text-based machine translation (MT) data to ST data via text-to-speech (TTS) systems.Yet, using TTS systems can be tedious and slow. In this work, we propose \textit{SpokenVocab}, a simple, scalable and effective data augmentation technique to convert MT data to ST data on-the-fly. The idea is to retrieve and stitch audio snippets, corresponding to words in an MT sentence, from a spoken vocabulary bank. Our experiments on multiple language pairs show that stitched speech helps to improve translation quality by an average of 1.83 BLEU score, while performing equally well as TTS-generated speech in improving translation quality. We also showcase how \textit{SpokenVocab} can be applied in code-switching ST for which often no TTS systems exit.\footnote{Our code is available at \url{https://github.com/mingzi151/SpokenVocab}}


\end{abstract}

%% file: sections/1-introduction.tex
\section{Introduction}
\label{sec:introduction}

End-to-end (E2E) speech-to-text translation (ST) models require large amounts of data to train~\cite{sperber2020speech}. Despite the emerging ST datasets~\cite{cattoni2021must,wang2021covost}, their size is considerably smaller compared to text-based  machine translation (MT) data.~A common remedy to tackle the data scarcity issue is to leverage text-based MT data in training ST systems. Common approaches include multi-task learning~\cite{anastasopoulos2018tied,ye2021end}, transfer learning \& pretraining~\cite{bansal2019pre,wang2020curriculum} and knowledge distillation~\cite{inaguma2021source,tang2021improving}. 

A more straightforward alternative is to convert text-based MT data to ST via text-to-speech (TTS) synthesis engines \cite{pino2019harnessing, jia2019leveraging}. This method is less commonly used despite its simplicity and effectiveness,\footnote{Only one work out of 8 uses TTS to augment data in the IWSLT2022 offline speech translation track. 
} mainly due to practical reasons: (i) TTS models have slow inference time and may incur monetary costs; 
(ii) the conversion is required for each MT datasets.
Recently, \citet{lam2022sample} proposed to generate synthetic speech without using TTS models. However, their approach is based on real ST data, and thus cannot be extended to MT data. 

\begin{figure}[t]
 \centering
   \includegraphics[scale=0.35]{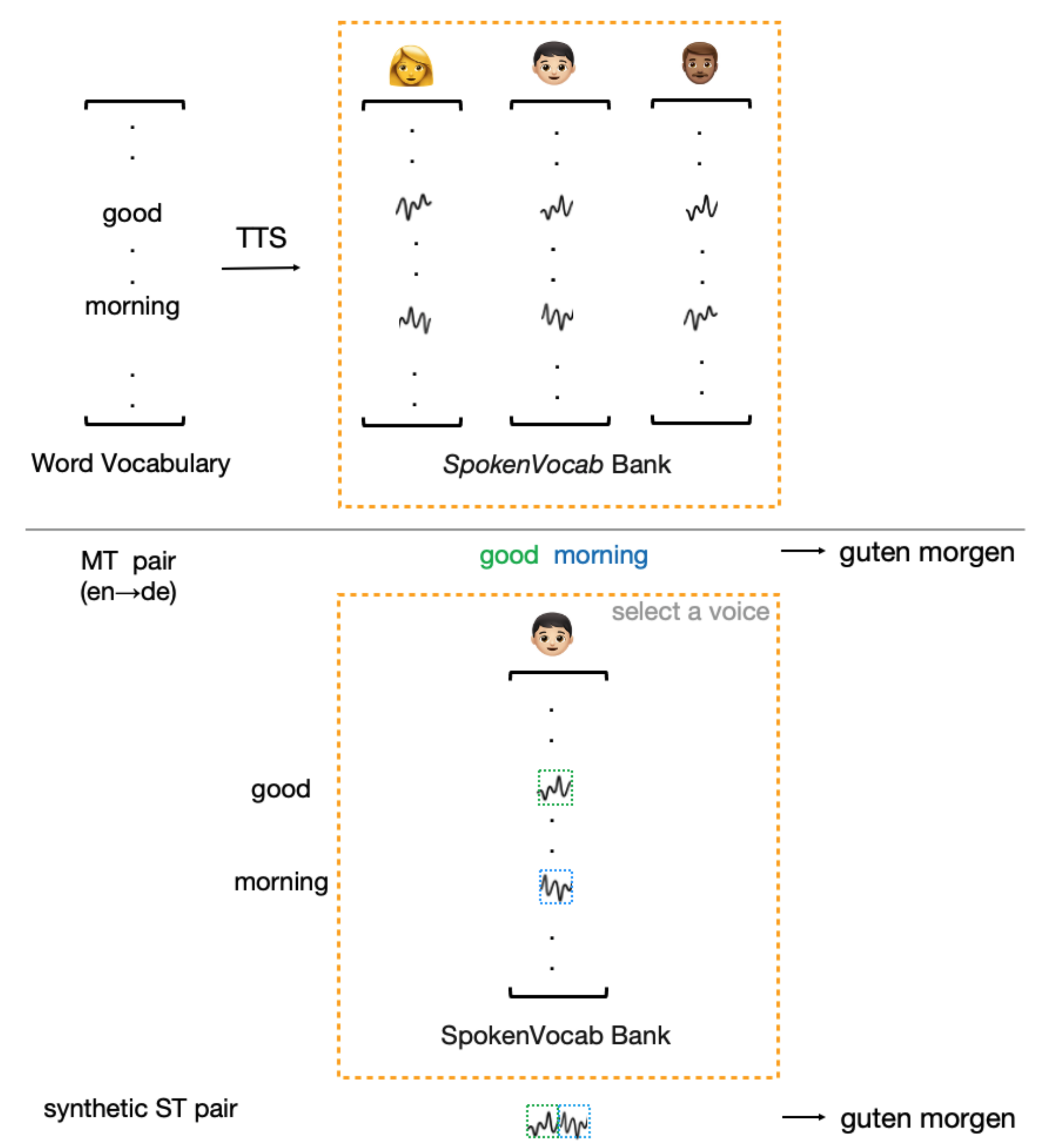}
 \caption{Overview of generating synthetic speech from SpokenVocab on-the-fly. 
 The first step is to prepare the SpokenVocab bank offline and the second step is to retrieve and stitch audio snippets from the bank by words in a sentence. 
 }
 \label{fig:method}
\end{figure}

In this work, we propose a simple, effective and efficient data augmentation approach to convert MT data to ST data on-the-fly. The idea is to prepare a set of spoken words, forming a spoken vocabulary~(\textit{SpokenVocab}) bank, and then generate synthetic speech by retrieving and stitching spoken words based on a text sequence, as shown in Figure \ref{fig:method}.\footnote{During the writing of this manuscript we found out that Voder, the first electronic speech synthesiser developed by Bell Labs in 1939, synthesized human speeches by decomposing it into its acoustic components and combining them using human operators in real time.} Our experiments show that this method is as effective as TTS-generated speech, at a much lower computational and financial cost. For instance, augmenting ST data on-the-fly with 100k of stitch-converted MT data, boosts translation quality by an average of 1.83 BLEU over 3 language pairs from {Must-C}~\cite{cattoni2021must} with no additional cost, memory, or speed footprints. Comparing the real ST data vs. our converted version from the same transcripts, to our positive surprise, revealed that our synthetic data outperforms its real counterpart by 0.41 BLEU score. We conduct thorough experiments to examine \textit{SpokenVocab} in  boosting translation and further showcase its use and benefit in the context of code-switching (CS) ST.  

We hope this simple technique to ease the use of MT data for ST in practice as well as other tasks where synthetic speech is useful.

%% file: sections/2-method.tex
\section{SpokenVocab}\label{method}

We describe our methodology in creating effective synthetic ST data based on MT data in this section. The core step is the preparation of a \textit{SpokenVocab} bank offline and stitching sounds on-the-fly. 

Concretely, 
we first use a TTS engine to convert items in a word vocabulary to speech to obtain a set of \textit{SpokenVocab} offline.\footnote{SpokenVocab could also be based on n-grams in a dataset.} Next, we can configure the TTS engine to generate 
different speaker voices and thus curate a \textit{SpokenVocab} bank in which each set corresponds to a "speaker". The purpose is to simulate, to the greatest extent, a realistic speech dataset consisting of various speakers. 
At training, assume we have access to an MT dataset 
and each pair denoted as~$<s, t>$~where $s$ and $t$ are source and targets sentences, respectively. Given such a pair, we choose one voice~\footnote{One could also generate utterances by mixing speakers at the token level, with no additional cost with our technique. We leave further investigation of this to future work as it requires a test condition (i.e., including various speaker voices per utterance) which is not available to the best of our knowledge.} 
from the bank, and produce synthetic speech by fetching corresponding audio snippets by words in $s$ from the bank and stitching them together. During stitching, we deploy cross-fade, a well-known technique to smooth transitions between two independent audio clips.\footnote{\url{https://github.com/jiaaro/pydub}} Pairing it with $t$ yields a synthetic ST instance.\footnote{We provide a \href{https://drive.google.com/drive/folders/17kJ8amwcGD6rgDuYaCdWFXmM5Kf01wGT?usp=sharing}{demo} for stitched speeches.}



%% file: sections/3-experiment.tex
\section{Experiments}
{We first present the ST system (\S\ref{st_model})  and TTS systems (\S\ref{tts_model})  used in this study. We then describe the ST and MT datasets (\S\ref{data}), followed by providing implementation details (\S\ref{impl}). Next we explain how \textit{SpokenVocab} is designed (\S\ref{prep}) and report translation results (\S\ref{main}). Lastly, we illustrate how our method can be applied to CS ST (\S\ref{cs}).} 
\subsection{Experimental Setup}\label{st_model}

\subsubsection{Speech Translation System} 
Pre-trained speech encoders and text decoders have shown great performance on ST \cite{li2020multilingual,zhao2022m}, compared to models trained from scratch. For this reason, we follow the architecture in \citet{gallego2021end} that uses Wav2vec~2~(W2V2) \cite{baevski2020wav2vec} as the speech encoder and mBart decoder \cite{liu2020multilingual} as the text decoder, joint with a lightweight linear adapter and a CNN-based length adapter. 

\subsubsection{TTS Systems} \label{tts_model}
To prepare \textit{SpokenVocab}, we use the Google TTS service,\footnote{\url{https://cloud.google.com/text-to-speech}} which supports a wide range of voice configurations; this allows simulating different speakers with various accents, gender and geographical background. We also use a off-the-shelf TTS toolkit, i.e., Tacotron2-DCA + Mulitband-Melgan (short for T2+Mel).\footnote{\url{https://github.com/mozilla/TTS}}  We use Google TTS to generate synthetic speech in  raw wavforms.




\subsubsection{Dataset} \label{data}
We conduct our major experiments on Must-C, a multilingual ST dataset curated from Ted talks. We focus on English (En)$\rightarrow$German (De), Romanian (Ro) and Italian (It). For MT data, we use a subset of WMT14, WMT16 and OPUS100\footnote{\url{http://opus.nlpl.eu/opus-100.php}} for De, Ro and It, with 100k, 100k and 24k instances, respectively.
%
%
For the code-switching (CS) setting, we use Prabhupadavani~\cite{sandhan2022prabhupadavani}, multilingual CS ST dataset, and we focus on En$\rightarrow$De, It. Its source utterances are code-mixed with English (major language), Bengali and Sanskrit; each utterance is translated manually to 25 languages. We prepare ST data following the instructions in~\citet{gallego2021end}. We preprocess MT data with the fairseq instructions and remove pairs with the length of target sentences greater than 64 words to avoid out-of-memory issues. Minimal preprocessing is performed on the CS ST dataset.
\subsubsection{Implementation Details} \label{impl}

Similar to \citet{li2020multilingual} and \citet{gallego2021end}, training different components of W2V2 and mBart decoder yields divergent results. 
In our initial experiments, we note that fine-tuning the entire W2V2 except for its feature extractor and freezing mBart lead to decent translation results, and thus we use this configuration for all our experiments. To ensure Must-C to be dominant, we make the ratio of Must-C and MT data to be approximately 8:1, unless mentioned otherwise. We use sacreBLEU~\cite{post2018call} to evaluate translation. Please refer to~\emph{Appendix} \ref{appendix:training details} for full training details, hyper-parameters and hardware. 



%% file: sections/4-results.tex
\subsection{SpokenVocab Preparation and Variations} \label{prep}

Constructing the \textit{SpokenVocab} bank is crucial, as synthetic speech produced in this manner have a direct impact on translation quality. In this section we examine \textit{SpokenVocab} from various dimensions. 

\stitle{TTS Conversion.}
The first questions to ask are which TTS system should be used to convert a word to a spoken form and what sampling rate (SR)
is appropriate.\footnote{SR is defined as the number of samples taken from a continuous signal per second.} To answer these questions, we conduct intrinsic evaluation on stitched speech by varying TTS engines and SR. Furthermore, as it is common to diversify raw wave forms with audio effects~\cite{potapczyk2019samsung}, we apply the same technique to distort our stitched speech. Results in Table \ref{tab:prep} show that using Google TTS and setting the SR to 24k are better choices, while distortion (i.e., adding the effects of tempo, speed and echo) may or may not be helpful. Contrary to the common practice of using a SR of 16k~\cite{baevski2020wav2vec}, applying 16k to \textit{SpokenVocab} alters the sound significantly, as shown in the demo in \S\ref{method}, and this has negative impacts on the system. Overall, we use the setting in \textit{italic} for the rest of our experiments.

\begin{table}[t]
\centering
\small
  \scalebox{1}
  {
  \begin{tabular}{lccccc}
    \toprule
       Data & TTS  & SR & Distort. & BLEU \\
    \midrule
     ST  & - & - & - &26.91 \\
    \hhline{======}
     \multirow{6}{7em}{\small{ST+MT$_\texttt{stitched}$}} &  T2+Mel                                              &  - &  - &  OOM  \\\cline{2-5}
     & \multirow{4}{3em}{\textit{Google}}                                            & \textit{24k} & \textit{-} &  \textbf{28.02}\\
    & &24k & \checkmark& 27.72\\\cline{3-5}
    & &16k & - &  26.77\\
    & &16k & \checkmark& 27.47\\
    \bottomrule
  \end{tabular}}
  \caption{Comparison of different TTS conversions in terms of TTS engine, sampling rate (SR) and distortion (Distort.) Top row: baseline. Bottom rows: MT data is converted to ST data with \textit{SpokenVocab}. OOM: out-of-memory with 24k and 16k SRs. \textit{italic}: best setting.
  }
  \label{tab:prep} 
\end{table}

\stitle{Word Vocabulary.} 
We compile a word vocabulary, consisting of 1) a common subset of  words\footnote{ The list comes from Official Scrabble Players Dictionary and Wiktionary's word frequency lists, and can be found at \url{https://github.com/dolph/dictionary/blob/master/popular.txt}}, and 2) unique words with a frequency of higher than 99 from the 
En$\rightarrow$X WMT subset. The purpose is to construct an approximated version of \textit{SpokenVocab} that is ready to convert any sentence to synthetic speech.  
For words that are not covered by the list, we employ a fuzzy matching mechanism where the most similar word at the surface level is returned. For instance, an out-of-vocabulary (OOV) word "apples" is replaced by its closest match in the vocabulary "apple", and the speech snippet for "apple" is retrieved. When no match is found, a default filter word, "a", is returned. To investigate the effect of this approximation which would inevitably lead to mispronounced words, we prepare another set of \textit{SpokenVocab} containing the full set of spoken words in the WMT data (eliminating the need for fuzzy matching). In controlled experiments on En$\rightarrow$De, 
the BLEU scores with the approximated and full \textit{SpokenVocab}s, with the size of 35k and 460k respectively, are 28.02 and 27.91. The negligible difference indicates the effectiveness of using an approximated \textit{SpokenVocab}. Additional ablation studies on using 50\% and 10\% of the full vocabulary yield scores of 27.79 and 27.94, further validating the insensitivity of \textsc{w2v2} to nuanced mispronunciation, perhaps due to the presence of powerful pre-trained auto-regressive decoder.\footnote{{Optionally, one can dynamically call a TTS system to generate an audio on OOV words.}}       

\stitle{Number of Speakers.}
Despite the artificial nature of the stitched speech sounds, one still can tell the speaker's information (e.g., gender, accent). To examine whether diverse voices would be helpful for translation, we set $n$ to 1, 5 and 10 and train models with the same amount of data. These systems display similar translation performance with 28.02, 27.73 and 27.80 BLEU scores respectively, suggesting that having a single speaker is sufficient. Our conjecture to this phenomenon is that speech representations produced by \textsc{w2v2} have removed speaker information, as demonstrated in~\citet{nguyen2020investigating} where analysis was conducted on wav2vec~\cite{schneider2019wav2vec}, the predecessor to \textsc{w2v2}. This could be further examined with using dialect- or pronunciation-focused translation settings, which we leave to future work.

\begin{table}[t]
\centering
\small
  \scalebox{0.9}
  {
  \begin{tabular}{lcccccc}
    \toprule
      & \multicolumn{3}{c}{Cost}  & \multicolumn{3}{c}{BLEU} \\
      \cmidrule(lr){2-4}\cmidrule(lr){5-7}
      Data & \faClockO & \faDollar & \faDatabase  & De & Ro & It \\
     \midrule
     \addlinespace[0.3em]
     ST &   & - &  & 26.91 & 24.66 & 22.13 \\
    \midrule
    \addlinespace[0.3em]
     ST + MT$_\texttt{TTS}$&  900 & 90 & 25 & \textbf{28.20} & 24.71 & \textbf{26.46} \\
    \midrule
     \addlinespace[0.3em]
     ST + MT$_\texttt{stitched}$ & 9 & 0 & 0 & 28.02 & \textbf{25.05} & 26.13  \\
     \addlinespace[0.3em]
    \bottomrule
  \end{tabular}}
  \caption{Translation quality on Must-C and the average costs  associated for generating synthetic speech for \textit{every 100k sentences} in terms of inference time in minutes (\faClockO), USD value (\faDollar) and storage required in GB~(\faDatabase). Preparing \textit{SpokenVocab} took 2 hours, free of charge, with Google TTS, and stitched speeches are discarded.}
  \label{tab:focus} 
\end{table}


\subsection{Translation Performance on Must-C} \label{main}
Producing synthetic speech from \textit{SpokenVocab} on-the-fly makes the conversion from text to speech highly scalable in terms of time and monetary costs, and it also avoids the need of storing speech. Table~\ref{tab:focus} reports the time, dollar value and space required to produce every 100k speech with Google TTS, while these numbers are negligible for \textit{SpokenVocab} due to its re-usability.\footnote{For fair comparison between TTS which operates on the full vocabulary, we report the cost under the full vocabulary version of our method.} Apart from scalability, it is more important to see the translation performance difference between unnatural  speech produced by \textit{SpokenVocab} and fluent speech generated by state-of-the-art TTS systems. Table~\ref{tab:focus} summarises results for 3 Must-C language pairs, with stitched speech and TTS-generated speech. As expected data augmentation of ST with MT data method boosts translation quality, using our method by 1.83 BLEU score on average. Our stitched speech performs equally well as TTS-generated counterpart, showing no loss of quality during conversion.

\subsection{Stitched Speech vs. Real Speech} \label{synt-real}
An alternative approach to augmentation is to leverage real ST data from any other existing domains. To assess whether our approach as another augmentation technique is still competitive, we conduct an experiment on En$\rightarrow$De by augmenting Must-C with 35k training instances from the Europarl-ST~\cite{iranzo2020europarl}. Table~\ref{tab:real-stit} reports the results. To our positive surprise, our stitched speech (generated from the transcripts of eurorparl-ST counterpart) works even better than the real Europarl-ST speech.  

\begin{table}[t]
\centering
\small
  {
  \begin{tabular}{llc}
    \toprule
     Data & Nature of Speech  & BLEU \\
     \midrule
     \addlinespace[0.3em]
     Must-C & real  & 26.91  \\
     \addlinespace[0.3em]
     Must-C + Europarl & real + real   & 27.5 \\
     \addlinespace[0.3em]
     Must-C + Europarl$_\texttt{TTS}$ & real + synthetic   & 27.76  \\
     \addlinespace[0.3em]
     Must-C + Europarl$_\texttt{stitched}$ & real + synthetic    & \textbf{27.91}  \\
     \addlinespace[0.3em]
    \bottomrule
  \end{tabular}}
  \caption{BLEU scores under different augmentations.}
  \label{tab:real-stit} 
\end{table}




\subsection{Code-switching Speech Translation} \label{cs}
Development in CS ST is constrained by the availability of relevant datasets~\cite{sandhan2022prabhupadavani} and using TTS systems to augment data is practically difficult.~To this end, our method provides a high degree of flexibility in that it can stitch audio clips of different languages freely.~To produce a code-switched utterance, we further prepare \textit{SpokenVocab} for Bengali (Google TTS does not support Sanskrit) based on an English-Bengali dictionary.\footnote{\url{https://github.com/MinhasKamal/BengaliDictionary}} 
We maintained the ratio of code-switching in the real data (i.e., 0.35 probability of CS occurring, and 2 as the average number of code-switched words in a sentence).~Please see Algorithm \ref{alg:cs} in \emph{Appendix} \ref{appendix:alg} for the detailed utterance generation process. Results in Table \ref{tab:cs} suggest that the models trained with additional 100k and 24k instances (for De and It respectively.)
from \textit{SpokenVocab} outperform those only trained with the original data. 

\begin{table}[t]
\centering
\small
  \scalebox{0.9}
  {
  \begin{tabular}{llcc}
    \toprule
    &&ST$_{\texttt{CS}}$ & ST$_{\texttt{CS}}$+MT$_{\texttt{CS-stitched}}$\\
    \cmidrule(lr){3-4}
    \multirow{2}{*}{BLEU}&En-Be$\rightarrow$De &26.11&28.09\\
    &En-Be$\rightarrow$It&26.41&26.90\\
    \bottomrule
  \end{tabular}}
  \caption{Translation quality for CS ST dataset. 
  }
  \label{tab:cs} 
\end{table}

%% file: sections/5-conclusion.tex
\section{Conclusion}
\label{sec:conclusion}
In this work, we proposed a simple, fast and effective data augmentation technique, \textit{SpokenVocab} for ST. This provides an alternative for converting MT data to ST data with TTS systems which comes with monetary and computation costs in practice. Our approach generates synthetic speech on-the-fly during training, with no cost or footprint.
We have shown that speech stitched from \textit{SpokenVocab} works as effective as TTS-generated speech, and unlike TTS system, it could directly be applied as a data augmentation tool in code-switching ST. Our approach can be used in other content-driven speech processing tasks as an uncompromising and easy-to-use augmentation technique.

%% file: sections/limitation.tex
\section*{Limitations}
CS ST exbihit difficulties \cite{huber2022code, weller2022end}, exposing several limitations in this study: 1) Bengali and Sanskrit (another minority language) are treated without difference, as they originate from the same script and Sanskrit is not supported by the Google TTS service. 2) We use a open-source language detection tool to calculate the oracle hyper-parameters in the dev set; yet, imperfection of the detector on token-level prediction and the fact that source sentences are written in Latin regardless of the language deviate the scores from true values.

%% file: sections/6-appendix.tex
\label{sec:appendix}



\subsection{Implementation Details}\label{appendix:training details}

We implement and train all models with fairseq\footnote{\url{https://github.com/facebookresearch/fairseq}} on 4 A40 GPUs, using 16 floating point precision, for $25k$ updates. \textsc{Wav2vec 2}\footnote{\url{https://dl.fbaipublicfiles.com/fairseq/wav2vec/wav2vec\_vox\_960h\_pl.pt}} and the mBart50\footnote{\url{https://dl.fbaipublicfiles.com/fairseq/models/mbart50/mbart50.ft.1n.tar.gz}}  decoder are used.  We employ an Adam optimizer with $\beta_1 = 0.99$,  $\beta_2 = 0.98$, while setting the dropout to 0.1, clip norm to 20 and label smoothing to 0.2. For the baseline models, we use a learning rate of 5e-04 and reduce it at plateau. For models trained with additional data, we use the same learning scheduler with a learning rate of 3e-04.


\subsection{Code-switching Speech Translation}\label{appendix:alg}

\renewcommand{\algorithmicensure}{\textbf{Output:}}
\begin{algorithm}[h]
\caption{Code-switching Utterance Generation}
\begin{algorithmic}[1]
\REQUIRE{$E,B:$ English and Bengali SpokenVocab, $Dict:$ English-Bengali Dictionary, $Keys:$ English words in $Dict$, $X:$ English sequence, $p:$ probability of cs  occurring, $n:$ number of code-switched words, $FetchSpeech:$ function to fetch speech}
\ENSURE{$U$: CS utterance}\\
\STATE $q = \text{NormDist}(0, 1)$
\IF{$q > p $} 
\STATE // Select words to be code-switched
\STATE  $words, indices = \text{Random}(X, n)$
\FOR {$word, i \text{ in } words, indices$}
\STATE // Only switch words in the dictionary 
\IF{$word \text{ in } Keys$}
\STATE // Replace with the Bengali word
$X[i] = Dict[word]$
\ENDIF
\ENDFOR
\ENDIF
\STATE $U = \text{FetchSpeech}(E, B, X)$
\RETURN $U$
\end{algorithmic}
\label{alg:cs}
\end{algorithm}